\documentclass[sigconf]{acmart}

\copyrightyear{2020}
\acmYear{2020}
\setcopyright{acmcopyright}
\acmConference[SIGIR '20] {43rd International ACM SIGIR Conference on Research and Development in Information Retrieval}{July 25--30, 2020}{Virtual Event, China}
\acmBooktitle{43rd International ACM SIGIR Conference on Research and Development in Information Retrieval (SIGIR '20), July 25--30, 2020, Virtual Event, China}
\acmPrice{15.00}
\acmDOI{10.1145/11221.27}
\acmISBN{978-1-4503-8016-4/20/07}
\usepackage{multirow}
\usepackage{algorithm}
\usepackage{algorithmic}
\usepackage{amsmath}
\usepackage{color}
\renewcommand{\algorithmicrequire}{\textbf{Input:}}  % Use Input in the format of Algorithm
\renewcommand{\algorithmicensure}{\textbf{Output:}} % Use Output in the format of Algorithm

\newcommand{\method}{\textbf{{ACKRec}~}}

\newcommand{\nop}[1]{}

\hypersetup{draft}

\begin{document}

\fancyhead{}
\title[Attentional GCNs for Knowledge Concept Recommendation]{Attentional Graph Convolutional Networks for Knowledge Concept Recommendation in MOOCs in a Heterogeneous View}

\author{Jibing Gong}
\authornote{The first two authors contributed equally.}
\affiliation{\institution{Yanshan University}}
\email{gongjibing@ysu.edu.cn}

\author{Shen Wang}
\authornotemark[1]
\affiliation{\institution{University of Illinois at Chicago}}
\email{swang224@uic.edu}

\author{Jinlong Wang}
\affiliation{\institution{Yanshan University}}
\email{wangjinlong@stumail.ysu.edu.cn}

\author{Wenzheng Feng}
\affiliation{\institution{Tsinghua University}}
\email{fwz17@mails.tsinghua.edu.cn}

\author{Hao Peng}
\affiliation{\institution{Beihang University}}
\email{penghao@act.buaa.edu.cn}

% \author{Dan Wang}
% \author{Yi Zhao}
% \author{Huanhuan Li}
% \affiliation{\institution{Yanshan University}}

\author{Jie Tang}
\authornote{Corresponding author.}
\affiliation{\institution{Tsinghua University}}
\email{jietang@tsinghua.edu.cn}

\author{Philip S. Yu}
\affiliation{\institution{University of Illinois at Chicago}}
\email{psyu@uic.edu}

\renewcommand{\shortauthors}{Wang and Gong, et al.}

\begin{abstract}

Massive open online courses (MOOCs) are becoming a modish way for education, which provides a large-scale and open-access learning opportunity for students to grasp the knowledge. To attract students' interest, the recommendation system is applied by MOOCs providers to recommend courses to students. However, as a course usually consists of a number of video lectures, with each one covering some specific knowledge concepts, directly recommending courses overlook students' interest to some specific knowledge concepts. To fill this gap, in this paper, we study the problem of knowledge concept recommendation.
We propose an end-to-end graph neural network based approach called \textit{\underline{A}ttentional Heterogeneous Graph \underline{C}onvolutional Deep \underline{K}nowledge \underline{Rec}ommender} (\textbf{ACKRec}) for knowledge concept recommendation in MOOCs. 
Like other recommendation problems, it suffers from sparsity issue. To address this issue, we leverage both content information and context information to learn the representation of entities via graph convolution network.
In addition to students and knowledge concepts, we consider other types of entities (e.g., courses, videos, teachers) and construct a heterogeneous information network (HIN) to capture the corresponding fruitful semantic relationships among different types of entities and incorporate them into the representation learning process. Specifically, we use meta-path on the HIN to guide the propagation of students' preferences. With the help of these meta-paths, the students' preference distribution with respect to a candidate knowledge concept can be captured. Furthermore, we propose an attention mechanism to adaptively fuse the context information from different meta-paths, in order to capture the different interests of different students.
To learn the parameters of the proposed model, we propose to utilize extended matrix factorization (MF). 
A series of experiments are conducted, demonstrating the effectiveness of \method across multiple popular metrics compared with state-of-the-art baseline methods.
%\nop{to compare with various baseline methods.}
The promising results show that the proposed \method is able to effectively recommend knowledge concepts to students pursuing online learning in MOOCs.

\end{abstract}

% <concept_id>10002951.10003317.10003347.10003350</concept_id>
% <concept_desc>Information systems~Recommender systems</concept_desc>
% <concept_id>10002951.10003260.10003261.10003269</concept_id>
% <concept_desc>Information systems~Collaborative filtering</concept_desc>
% <concept_significance>300</concept_significance>
% <concept_id>10002951.10003260.10003261.10003271</concept_id>
% <concept_desc>Information systems~Personalization</concept_desc>
% <concept_significance>300</concept_significance>
% <concept_id>10002951.10003260.10003261.10003270</concept_id>
% <concept_desc>Information systems~Social recommendation</concept_desc>
% <concept_significance>100</concept_significance>
% <concept_id>10010147.10010257.10010293.10010294</concept_id>
% <concept_desc>Computing methodologies~Neural networks</concept_desc>




\maketitle


%   \includegraphics[width=\textwidth]{images/framework.pdf}
%   \label{fig:framework}

\section{Introduction}

In recent years, massive open online courses (MOOCs) are gradually becoming a mode of alternative education worldwide. 
For example, Coursera, edX, and Udacity, the three pioneering MOOC platforms, offer millions of user accesses to numerous courses from internationally renowned universities. 
In China, millions of users study in \textit{XuetangX}\footnote{http://www.xuetangx.com}, which is one of the largest MOOC platforms\cite{ModelingandPredicting}, where thousands of courses are offered on various subjects. 
Although the number of students in MOOCs is continuously growing, there are still some straits with MOOCs. 
A challenging problem for MOOCs is how to attract students to study continuously and efficiently on the platforms, where the overall course completion rate is lower than 5\% \cite{SmartJump}. Therefore, it requires better understanding and capturing of student interests.
\nop{One reason is that these platforms lack of in-depth understanding of student interests and cannot help users build a complete knowledge concept system.}

% \shen{second paragraph need to indicate the importance of proposing new problem: knowledge recommendation}
%MOOCs \nop{have transformed traditional learning methods and} provide an opportunity to research students' online learning behavior \cite{GuessYouLike}. 
To understand and capture student interests on MOOCs platforms, multiple efforts have been done, including course recommendation \cite{GuessYouLike,zhang2019hierarchical}, behavior prediction \cite{ModelingandPredicting}, user intentions understanding \cite{SmartJump}, etc. Among these efforts, recommendation system is applied by MOOCs provider to recommend courses to students. However, a course usually consists of a number of video lectures with each one covering some specific knowledge concepts. Direct course recommendation overlooks students' interest to specific knowledge concept, e.g., computer vision courses taught by different instructors may be quite different in a microscopic view (cover different sets of knowledge concepts): someone instructor may only cover geometry based methods while other one may only cover deep learning based methods, and thus recommending the computer vision course only covering geometry based methods to the student interested in the deep learning based methods will not be a good match.
%However, different students may have different interests and different background
%In fact, the course recommendations help users choose courses, but it cannot provide further assistance during the learning process. 
%\nop{The models in these systems only consider relationships between single types of entities while ignoring the feature structures connecting different types of entities.} 
Therefore, it requires to study students' online learning interests from a microscopic view and conduct knowledge concept recommendation.
%We model the problem as a personalized knowledge concept recommendation system for MOOCs.  The knowledge concepts have a series of labels, which belong to courses and videos. When learning in MOOCs, knowledge concept recommendation can help students review and consolidate the learned knowledge concepts, and reduce the difficulty of learning new knowledge concepts.

%according to the relationship between different knowledge concepts of different courses
%When learning a new course, the knowledge concept recommendation not only provides an opportunity for those students who have some basic knowledge to review and consolidate, but also reduce the difficulty of learning an unfamiliar course for the students without knowledge reserve.

\begin{figure}
\centering
\includegraphics[width=8cm]{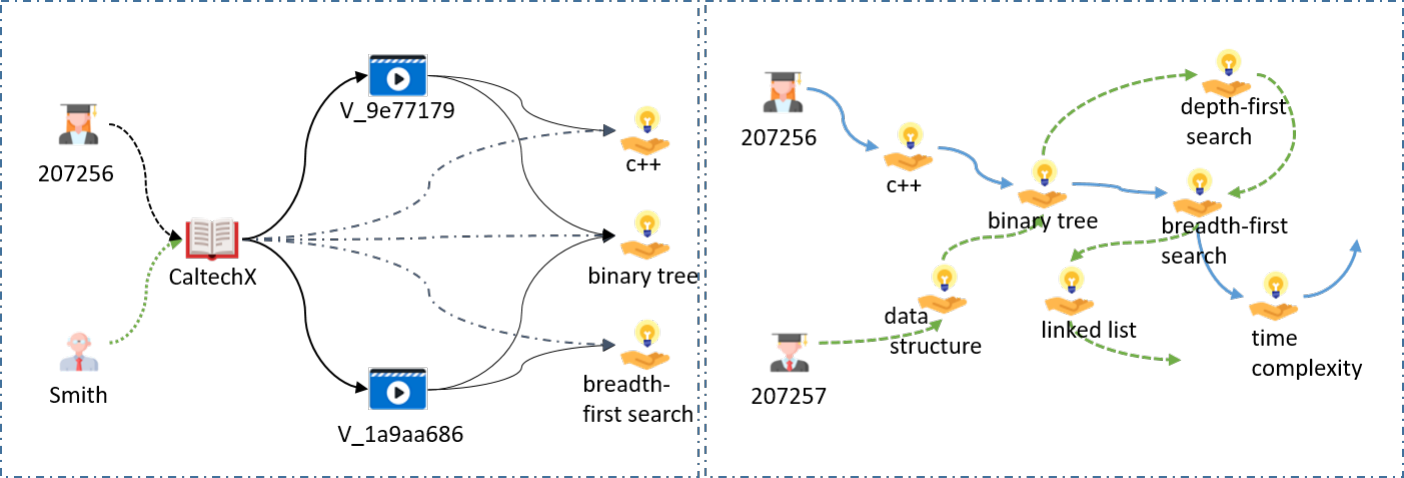}  
\vspace{-5pt}
\caption{A comprehensive view of dataset collected from MOOCs. The left is the system structure of MOOCs. The right is the structure rebuild by users' behaviors of online learning. Different types of lines indicate different relationships between pairs of entities.}
\label{fig:data}
\vspace{-12pt}
\end{figure}

Traditional recommendation strategy, such as collaborative filtering (CF), which considers user (students) historical interactions and makes recommendations based on potential common preferences from users with similar interests, has achieved great success. However, CF based methods suffer from the sparsity of user-item (student-knowledge concept) relationships, which limits the recommendation performance. To overcome this problem, a number of efforts have been done by leveraging side information, such as social networks \cite{jamali2010matrix}, user/item attributes \cite{wang2018shine},
images \cite{zhang2016collaborative}, contexts \cite{sun2017collaborative}, etc. 
In a MOOCs platform, we observe that in addition to the user and knowledge concept, there exist multiple types of entities (video, course, teacher) and multiple types of relationships between pair of different entities. 
Table \ref{tab:MOOCData} shows the statistics of the real-world XuetangX data collected between January 1st, 2018 and March 31st, 2018. 
This data consists of 9,986 users, 43,405 videos, 7,020 courses, 5,038 teachers, 1,029 knowledge concepts, and corresponding multiple types of relationships. 
As shown in Figure \ref{fig:data}, the ``\textit{course: V\_9e77179}'' includes the ``\textit{knowledge concept: c++}'', the ``\textit{student: 207256}'' is taking the ``\textit{course: CaltechX}'', the ``\textit{video: V\_1a9aa686}'' is related to the ``\textit{knowledge concept: binary tree}'', and the ``\textit{course: CaltechX}'' is taught by the ``\textit{teacher: Smith}''. 
Further more, taking users' behavior history into consideration, we can discover additional relationships. For example, the ``\textit{user: 207256}'' clicked the ``\textit{knowledge concept: c++}'', ``\textit{knowledge concept: binary tree}'', and ``\textit{knowledge concept: depth-first search}''. 
Accounting for above multiple types of relationships, 
%\nop{and construct a heterogeneous information network (HIN) \ cite{}, }
we can get much more fruitful facts and interactions between the user and knowledge concepts.
If we merely depend on the basic structures, it is difficult to find the significant interaction between ``\textit{knowledge concept: depth-first search}'' and ``\textit{knowledge concept: time complexity} '', which belong to different courses but are clicked by one user. As shown in Figure \ref{fig:data}, different knowledge concepts contain different context. 
%Further, different knowledge concepts have different relationships. 
Only utilizing single type of interaction may overlook the significant relations between user and knowledge concept.
For example, ``\textit{knowledge concept: c++}'' and ``\textit{knowledge concept: binary tree}'' have dissimilar semantics even though they are included in the same video. 
These heterogeneous relationships provide rich side information and can benefit the recommendation system in three folds: (1) semantic relatedness among knowledge concepts can be introduced and help to identify the latent interaction; (2) a user's interests can be reasonably extended and the  diversity of recommended knowledge concepts can be increased; and (3) a user's interest can be interpreted by tracking a user's historical records along these relationships.
Thus, it requires to incorporate these heterogeneous relationships into the representation learning of the entities.

\begin{table}
\caption{Statistics of entities and relations for dataset.}
\vspace{-5pt}
\label{tab:MOOCData}
\begin{tabular}{cc|cl}
  \toprule
  Entities & Statistic & Relations & Statistic \\
  \midrule
  user & 9,986 & user-course & 14,326 \\
  video & 43,405 & course-video & 87,129 \\
  course & 7,020  & teacher-course & 13,274 \\
  teacher & 5,038  & video-knowledge concept & 11,732 \\
  knowledge & \multirow{2}{*}{1,029} & \multirow{2}{*}{course-knowledge concept} & \multirow{2}{*}{21,507} \\
  concept & ~ & ~ & ~ \\
\bottomrule
\end{tabular}
\vspace{-7pt}
\end{table}

Based on above observation, we propose \textit{\underline{A}ttentional Heterogeneous Graph \underline{C}onvolutional Deep \underline{K}nowledge \underline{Rec}ommender} (\method), an end-to-end framework for knowledge concept recommendation on MOOCs platform. 
To capture heterogeneous complex relationships, we model the MOOCs platform data as a heterogeneous information network (HIN) \cite{Heterogeneous}. 
Then, we propose an attention-based graph convolutional networks (GCNs) to learn the representation of different entities. Traditional GCNs can only capture the homogeneous relationships among homogeneous entities, which overlooks the rich information among heterogeneous relationships. To address this issue, we use meta-paths \cite{sun2017collaborative} as the guidance to capture the heterogeneous context information in a HIN via GCN. In this way, the heterogeneous relationships are utilized in a more natural and intuitive way. Moreover, considering that different students may have different interests, we further propose an attention mechanism to adaptively leverage context in multiple meta-paths. In the end, we propose to optimize the parameters of proposed model via an extended matrix factorization and obtain the final recommendation list.
The key contributions of this paper can be summarized as follows:
% The key contributions of this paper can be summarized as follows:
\begin{itemize}
\vspace{-8pt}
  \item We identify the important problem of knowledge concept recommendation, which is often overlooked by the existing MOOCs recommendation system. Knowledge concept recommendation fills this gap and provides a more microscope level recommendation.
  %To the best of our knowledge, this is the first attempt to recommend knowledge concepts to students during online learning in MOOCs. We pay more attention to improve the effectiveness of students online learning in MOOCs from a more subtle and deeper perspective. The knowledge concepts recommender system can give some suggestions about knowledge concepts, which is a new type of tag to describe the course and video.
  \nop{It can be regarded as an extension of course recommendation in a more detailed view.}
  %\shen{The difference is not clear. ``how to improve the effectiveness of students online learning in MOOCs from a more subtle and deeper perspective`` this purpose has no clear metric to evaluate and prove}
  %When learning in MOOCs, knowledge concept recommendation can help students review and consolidate the learned knowledge, and reduce the difficulty of learning new knowledge according to the relationship between different knowledge concepts of different courses.
  
  \item We propose $\method$, a novel end-to-end framework utilizing rich heterogeneous context side information to assist knowledge concept recommendation.
  
  %which use rich heterogeneous context side information. leverages both content and rich heterogeneous context side information and propagating users' interest under the guide of meta-path.
  
  %automatically. discovers users’ hierarchical potential interests by iteratively propagating users’ preferences in the KG
  %Benefit from the attention mechanism and the powerful graph convolutional networks on the graph, we can obtain one low-dimensional vector to represent a user or a knowledge concept.  Then, we utilize matrix factorization to complete the knowledge concept recommendation task. Finally, all parameters are trained in an extended matrix factorization model. 
  %We integrate relation features and semantic features as the inputs for the representation of knowledge concepts.
  
  \item We develop a heterogeneous information network modeling to capture various complex interactions among different types of entities in the MOOCs platform.
  %We utilize a heterogeneous information network to model representations of different types of entities. The MOOC data is regarded as a heterogeneous information network, which involves multiple types of entities. The network not only incorporates the basic structure, but also the behaviors of all users. Instead of only using single features to describe the given knowledge concepts, we use semantic features and relation features to jointly perform the recommendation task of knowledge concepts.
  \item We design an attention-based graph convolutional network, which can incorporate both content and heterogeneous context together into the representation learning of different entities. The proposed model can automatically discovers user potential interests by propagating users' preferences under the guide of meta-path in an attentional way.
  
  \item We conduct numerous experimental studies using real-world data collected from XuetangX to fully evaluate the performance of the proposed model. We study the parameters, including meta-path combination, representation dimension, number of latent factories, and number of GCNs layers. We synthetically demonstrate the effectiveness of the proposed model compared with a series of strong baselines.
\end{itemize}
%\vspace{-5pt}
% The remainder of this paper is organized as follows. 
% Section~\ref{sec:system} introduces the system architecture. 
% Section~\ref{sec:method} describes the details of proposed method. 
% Section~\ref{sec:exper} discusses the experiments and analysis. 
% Section~\ref{sec:relatedwork} introduces the related works. 
% Finally, we provide our conclusions and future works in Section~\ref{sec:conclu}.

\vspace{-5pt}
\section{PROBLEM STATEMENT and SYSTEM ARCHITECTURE}
% \section{PROPOSED METHOD}

% \label{sec:system}
\subsection{Problem Statement}

Given a target user with corresponding interactive data \nop{when the user online learning }in MOOCs, the goal is to calculate the interest score about the user and a series of knowledge concepts, and the recommend results -- a top $N$ list of knowledge concepts. 
More formally, given interactive data of a user $u$, a predict function $f$ is learned and used to generate a recommend list of knowledge concepts $K$ (e.g., "c++", "binary tree", "linked list", ect.), such that $f: u \rightarrow \{k_i | k_i \in K, i < N\}$.

\begin{figure*}[t]
\centering
\includegraphics[width=\textwidth]{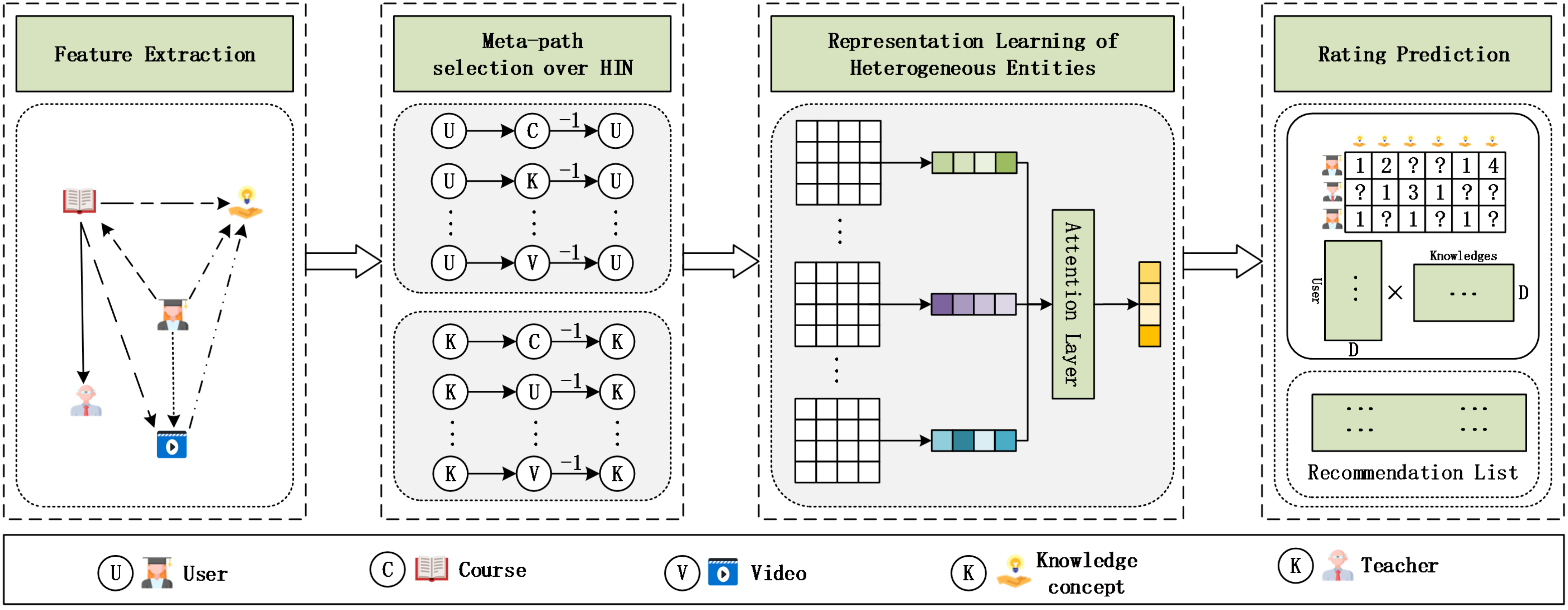}
\caption{System architecture of \method.}
\label{fig:system}
\end{figure*}
%  \textbf{Feature Extraction} part extract features from different types entities; \textbf{Meta-path Selection} part models the MOOCs data as a heterogeneous information network and describes the relationships of different entities; \textbf{Representation Learning of Different Entities} part utilizes attention-based graph convolutional networks to generate the representations of user and knowledge concept respectively; \textbf{Rating Prediction} part bases matrix factorization model to optimize parameters and obtain the final recommendation list.
\vspace{-5pt}
\subsection{System Architecture}

The architecture of our proposed knowledge concept recommendation system, \textbf{ACKRec}, is shown in Figure \ref{fig:system}. 
It consists of the following components:
\vspace{-5pt}
\begin{itemize}
  \item \textbf{\textit{Feature Extraction.}} By using the data collected from the MOOCs, we first extract content information as content feature from the knowledge concepts' name, and then analyze various relationships (e.g., $concept-video$ and $concept-course$ relations) among different types of entities (e.g., knowledge concept, video, course) to describe the knowledge concept. Similarly, we also generate the concept features and context feature for the user. (See Section 3.1 for details regarding feature extraction.)
  \item \textbf{\textit{Meta-path Selection.}} Based on the features extracted from the data, in this module we construct a structural HIN to model the relationships among different types of entities, and then select different meta-paths from the HIN to depict the relatedness over knowledge concept (i.e., with different meanings). For example, if two different users enrolled in the same course, we brings an edge between two users. (See Section 3.2 for details regarding the meta-path builder on HIN.)
  \item \textbf{\textit{Representation Learning of Heterogeneous Entities.}} Based on the meta-paths constructed in the previous step, a representation learning model is proposed to learn the low-dimensional representations of the entities in a heterogeneous view. The model is capable of capturing the structural correlations between heterogeneous entities. Specifically, we leverage the selected meta-paths to guide the entity representation learning via graph convolutional networks. Later, we utilize the attention mechanism to adaptively fuse the learned entity representations from different meta-paths. (See Section 3.3 for details regarding our proposed model \method.)
  \item \textbf{\textit{Rating Prediction.}} After generating the low-dimensional representations of users and knowledge concepts, the dense vectors of entities are fed to an extended matrix factorization to learn the parameters of the model. Moreover, we predict users' interests in the unclicked knowledge concepts base on the user-item (student-knowledge concept) rating matrix.
%   \shen{The user-item (student-knowledge concept) rating matrix contains the implicit feedback representing the users' online learning behaviors in the MOOCs.}
\end{itemize}
%\vspace{-5pt}

\vspace{-5pt}
\section{PROPOSED METHOD}
In this section, we introduce the details of how we learn the representation of knowledge concepts and users based on the generated content feature and context feature, and how we perform knowledge concept recommendation based on the learned representations.\label{sec:method}
\vspace{-18pt}
\subsection{Feature Extraction}
\subsubsection{Content Feature.}
In general, names of knowledge concepts are almost a generalization of knowledge concepts (e.g., ``\textit{c++},'' ``\textit{binary tree},'' ``\textit{linked list}''), which contains rich semantic information. Hence, we generate the word embedding of the name of the knowledge concept and use it as content feature for knowledge concept. Specifically, we use Word2vector \cite{mikolov2013efficient} to generate the word embedding. For the user concept, we generate the content feature in a similar way.  \nop{Besides the word embedding feature, Note that the content feature of knowledge concepts are studied specifically to facilitate the illustration of our proposed model.other types of features (e.g., bag-of-words, n-gram binaries, even one-hot) are also applicable. In fact, we use the one-hot code as the features of users.}
\vspace{-2pt}
\subsubsection{Context Feature.}
Content feature such as word embedding of knowledge concept names can be used to represent information of a knowledge concept. Besides, there exist rich context information, such as relationships between different entities in the network structure (e.g., \textit{user: 207256} watched video: \textit{v\_9e77179} and \textit{video: v\_1a9aa686}; this behavior implies a relation between two videos). To include these complex relationships among different types of entities, we further model the context information as the feature. 
Specifically, we consider the following relationships in a user learning activities.
\nop{The following samples illustrate relationships from user learning activities.}
\begin{itemize}
    \item \textbf{$R^{u}_{1}$}: Based on the data of users' online learning behaviors, we build the \textbf{\textit{user-click-knowledge concept}} matrix $\mathbf{A}_1^{u}$, where each element $c_{i,j} \in \{0,1\}$ implies that a user $i$ clicked a knowledge concept $j$ during his learning activities.
    \item \textbf{$R^{u}_{2}$}: To describe the relation between a user and a course, we generate the \textbf{\textit{user-learn-course}} matrix $\mathbf{A}_2^{u}$, where each element $l_{i,j} \in \{0,1\}$. It indicates that a user $i$ is taking a course $j$.
    \item \textbf{$R^{u}_{3}$}: Similarly, we generate the \textbf{\textit{user-watch-video}} matrix $\mathbf{A}_3^{u}$, where each element $w_{i,j} \in \{0,1\}$ denotes that a user $i$ has watched a video $j$.
    \item \textbf{$R^{u}_{4}$}: To describe the behavior that a user is taking a course, which is taught by a teacher, we generate the \textbf{\textit{user-learn-course-taught by-teacher}} matrix $\mathbf{A}_4^{u}$, where element $t_{i,j} \in \{0,1\}$ denotes that a user $i$ is taking a course taught by a teacher $j$.
\end{itemize}
We generate these relationship to describe the user related interactions in the heterogeneous information network.
\nop{These relationships are the user related interactions over heterogeneous information network.}
For knowledge concepts, we also discover a number of knowledge concepts related relationships.
% \begin{itemize}
%     \item \textbf{$R^{k}_{1}$}: The relationship between knowledge concept and course \textbf{\textit{knowledge concept-included by-course}} is described as a matrix $A_1^{k}$ where each element $I_{i,j} \in \{0,1\}$ indicates that a knowledge concept $i$ is included by a course $j$.
%     \item \textbf{$R^{k}_{2}$}: From the record of users' online learning behaviors, we construct the \textbf{\textit{knowledge concept-learnt by-user}} matrix $A_2^{k}$ where each element $l_{i,j} \in \{0,1\}$ denotes that a knowledge concept $i$ is learnt by a user $j$.
%     \item \textbf{$R^{k}_{3}$}: To show that a complex relationship among knowledge concept, course and teacher. We build the \textbf{\textit{knowledge concept-included by-course-taught by-teacher}} matrix $A_3^{k}$ where each element $l_{i,j} \in \{0,1\}$ denotes that a knowledge concept $i$ is mentioned in a class by a teacher $j$.
%     \item \textbf{$R^{k}_{4}$}: We generate the \textbf{\textit{knowledge concept-involved by-video}} matrix $A_4^k$ where each element $O_{i,j} \in \{0,1\}$ indicates knowledge concept $i$ is involved by a video $j$.
% \end{itemize}
For example, the relation \textbf{\textit{knowledge concept-included by-video}} denotes that a knowledge concept is included in a video, and the relation \textbf{\textit{knowledge concept-involved-course}} indicates that a knowledge concepts is covered in a course.
\vspace{-5pt}
\subsection{Meta-path Based Relationship}

To model different types of entities and their complex relationships in a proper manner, we first describe how to utilize a HIN to depict users, knowledge concepts, and corresponding heterogeneous relations among them.
\nop{in order to create knowledge concept recommendations for users.} 
% The concepts related to the HIN are as follows:
Before proceeding to our approach, we first introduce some related concepts.

Definition 1. \textit{\textbf{Heterogeneous information network (HIN)}} \cite{Heterogeneous}. A HIN is denoted as $\mathcal{G} = \{\mathcal{V}, \mathcal{E}\}$ consisting of an object set $\mathcal{V}$ and a link set $\mathcal{E}$. A HIN is also associated with an object type mapping function $\phi : \mathcal{V} \rightarrow \mathcal{N}$ and a link type mapping function $\varphi : \mathcal{E}\rightarrow \mathcal{R}$. $\mathcal{N}$ and $\mathcal{R}$ denote the sets of the predefined object and link types, where $|\mathcal{N}|+|\mathcal{R}|>2$.

%The HIN provides the structure of the data collected from the MOOCs, and generates abstractions of specific relations. 
In this study, we model the MOOCs data as the a HIN. Specifically, the constructed HIN includes five entities (i.e., user \textbf{(U)}, course \textbf{(C)}, video \textbf{(V)}, teacher \textbf{(T)}, knowledge concept \textbf{(K)} as shown in Figure \ref{fig:system}) and a series of relationships among them (e.g., \textbf{\textbf{$R^{u}_{1}$}, \textbf{$R^{u}_{2}$}, \textbf{$R^{u}_{3}$}, \textbf{$R^{u}_{4}$}}). Based on the constructed HIN, we can obtain the network schema, where their definitions are as followed.  

Definition 2. \textit{\textbf{Network schema}} \cite{wang2019attentional}. The network schema is denoted as $\mathcal{S}=(\mathcal{N}, \mathcal{R})$. It is a meta-template for an information network $\mathcal{G}=\{\mathcal{V}, \mathcal{E}\}$ with the object type mapping $\phi : \mathcal{V} \rightarrow \mathcal{N}$ and the link type mapping $\varphi : \mathcal{E} \rightarrow \mathcal{R}$, which is a directed graph defined over object types $\mathcal{N}$ with edges as relations from $\mathcal{R}$.

\begin{figure}
\centering
\includegraphics[width=8cm]{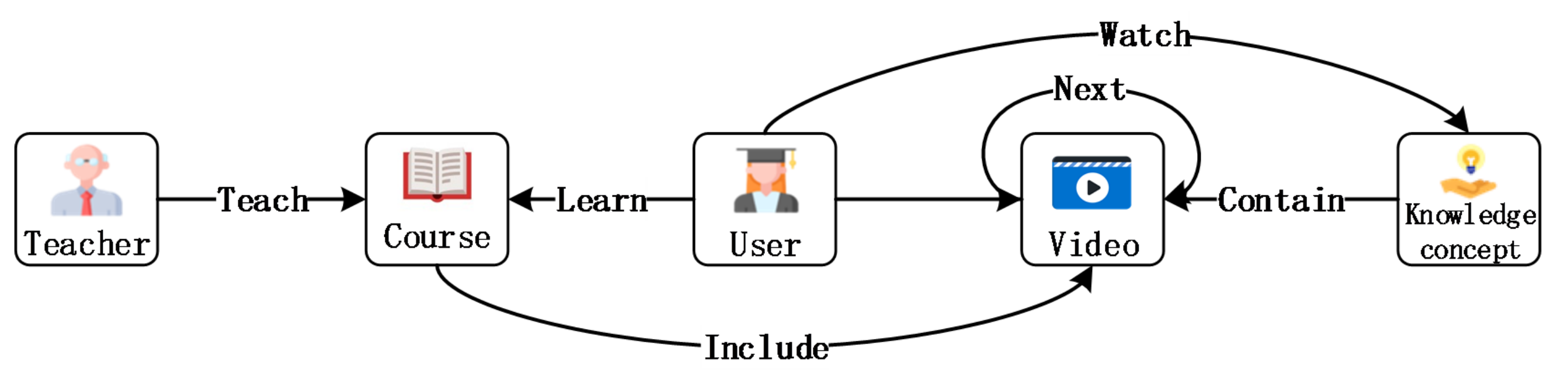}  
\caption{Network schema for HINs in MOOCs. The MOOCs include user, course, video, teacher and knowledge concept. Different types line indicate different type of relationships among different types entities.}
%  Different types line indicate different relationships among different types entities.
\label{fig:network}
\end{figure}

We defined our network schema in Figure \ref{fig:network}, which represents semantic and relation information comprehensively in the MOOCs dataset. Based on the the network schema, we can
discover the semantic paths between a pair of entities, which is called meta-path.

Definition 3. \textit{\textbf{Meta-path}} \cite{Anewmodel}. A meta-path is defined on a network schema $\mathcal{S}=(\mathcal{N}, \mathcal{R})$ and is denoted as a path in the form of $N_{1} \stackrel{R_{1}}{\rightarrow} N_{2} \stackrel{R_{2}}{\rightarrow} \cdots \stackrel{R_{l}}{\rightarrow} N_{l+1}$ (abbreviated as $N_{1}, N_{2}, \cdots N_{l+1}$), which describes a composite relation $\mathrm{R}=\mathrm{R}_{1} \circ R_{2} \cdots \circ R_{l}$ between object $N_{1}$ and $N_{l+1}$ , where $\circ$  denotes the composition operator on relations.

Typical meta-paths between two users can be defined as follows: $U \stackrel{\text { click }}{\longrightarrow} K \stackrel{\text { click }^{-1}}{\longrightarrow} U$, which means that two different users are related because they click the same knowledge concept; $U \stackrel{\text { learn }}{\longrightarrow} C \stackrel{\text { taught by }}{\longrightarrow} T \stackrel{\text { taught by }^{-1}}{\longrightarrow} C \stackrel{\text { learn }^{-1}}{\longrightarrow} U$, which denotes that two users are related through paths containing different courses taught by the same teacher. 
Notice that, the potential meta-paths induced from the HIN can be infinite, but not everyone is relevant and useful for the specific task of interest. Fortunately, there are some algorithms \cite{chen2017task} proposed recently for automatically selecting the meta-paths for particular tasks.
Given all the concepts about the HIN, we now proceed to our problem of Heterogeneous Information Network Representation Learning.
% Definition 4. \textit{\textbf{Heterogeneous Information Network Representation Learning}} \cite{metapath2vec}. This task is to learn a low dimensional latent representation $\mathbf{e}_i \in \mathbb{R}^{d}$, where $i$ denotes a entity in a heterogeneous information network $\mathcal{G}$ and $d << |\mathcal{V}| $. The learned $\mathbf{e}_i$ is expected to capture the complex structural and rich semantic relations among different entities in the heterogeneous information network $\mathcal{G}$.
% For example, the knowledge concept representation $\mathbf{e}^{k}_{i}$ is a low-dimensional vector, 
% \nop{and a knowledge concept $i$ is represented a $d$-dimensional vector in $i-th$ row of matrix $e^{k}$.}
% Given the HIN constructd from MOOCs data, we 
% The learned user representations and knowledge concept representations are able to apply to the knowledge concept recommendation task. In this problem, we propose to optimized the parameters via matrix factorization.
The notations we will use throughout the article are summarized in Table \ref{tab:notations}.

\begin{table}
\caption{Notations and explanations.}
\vspace{-5pt}
\label{tab:notations}
\begin{tabular}{c|cl}
    \toprule
    Notation&Explanation\\
    \midrule
    $\mathcal{G}$&heterogeneous information network\\
    \hline
    $\mathcal{V}$&set of entities\\
    \hline
    $\mathcal{E}$&set of relations\\
    \hline
    $\mathcal{S}$&network schema\\
    \hline
    $\mathcal{MP}$ &  set of meta-paths \\
    %\hline
    %$P$ & the number of meta-paths\\
    \hline
    \multirow{2}{*}{$\mathbf{A},\mathbf{\widetilde{D}}$} & adjacency matrix and degree matrix \\
    ~ & base different meta-paths \\
    \hline
    $\mathbf{X}$ & the features matrix of entities \\
    \hline
	$\mathbf{h}^{l}$ & $l-th$  layer of entity representation \\
	\hline
	$\mathbf{W}^{l}$ &  weights of $l-th$ GCN layer\\
	\hline
    $\mathbf{e}$ & the representation of entities \\
    \hline
    $\alpha$ & the weights of meta-paths \\
    \hline
    $\mathbf{x}_u$&the latent factors of user $u$ \\
    \hline
    $\mathbf{y}_k$&the latent factors of knowledge concept $k$ \\
    \hline
    \multirow{2}{*}{$m,n$} & the number of users and \\
    ~ & knowledge concepts in MOOCs dataset \\
    \hline
    $d$ & the representation dimension \\
    \hline
    $D$ &  the number of latent factors \\
    \hline
    $\mathbf{r}_{u,k}$ & the true rating of user $u$ to knowledge concept $k$ \\
    \hline
    $\mathbf{\widehat{r}}_{u,k}$ & predicted rating of user $u$ to knowledge concept $k$ \\
    \hline
    \multirow{2}{*}{$\mathbf{t}$} & the matrix to integrate the $\mathbf{e}^{u}$ and $\mathbf{e}^{k}$  \\
    ~ & be in the same space \\
    \hline
    $\beta$ & the tuning parameter \\
    \bottomrule
  \end{tabular}
\end{table}
\vspace{-5pt}
\vspace{-6pt}
\subsection{Attention-based Graph Convolutional Networks for HIN Representation Learning}

After the content features and context features are obtained, we feed the entity content features to the graph convolutional networks to learn the latent entity representation. 
%Here, we only discuss the representation learning process of knowledge concepts. The users representation is learned in a similar way. 
Given the heterogeneous information network $\mathcal{G}=(\mathcal{V}, \mathcal{E})$ associated with a set of meta-paths $\mathcal{MP}=\left\{MP_{1}, MP_{2}, \cdots MP_{\|MP\|} \right\}$ and the corresponding adjacency matrix $\mathcal{A}=\left\{\mathbf{A}_{1}, \mathbf{A}_{2}, \cdots \mathbf{A}_{|MP|}\right\}$. $|MP|$ denotes the number of meta-paths. We adopt a multiple-layer graph convolutional network (GCN) with the following layer-wise propagation rule:
\vspace{-5pt}
\begin{equation}
\mathbf{h}^{(l+1)}=\sigma\left(\mathbf{P} \mathbf{h}^{l} \mathbf{W}^{l}\right),
\end{equation}
\vspace{-2pt}
Here we remove the subscripts of meta-path indicator, user indicator and knowledge concept indicator for all the graph related symbols for simplification. $\mathbf{h}^{(l+1)}$ denotes the new representation of an entity. In particular, $\mathbf{h}^{0}$ is the content feature we have extracted at the first step. $\mathbf{P}=\mathbf{\widetilde{D}}^{-1 / 2} \mathbf{\widetilde{A}} \mathbf{\widetilde{\mathbf{D}}}^{-1 / 2}$, $\mathbf{\widetilde{A}}=\mathbf{A}+\mathbf{I}$ is the adjacency matrix corresponds to a specific meta-path with self-connections and $\mathbf{I}$ is the identity matrix, $\mathbf{\widetilde{D}}=\operatorname{diag}(\mathbf{\widetilde{A}} \mathbf{1})$, and $\mathbf{1}$ is the all-ones vector. Here $\sigma(\cdot)$ is defined as $R e L U(\cdot)$, where $ReLU(a) = \max{\{0,a\}}$ is an entry-wise rectified linear activation function. $l$ is the layer number indicator. $\mathbf{W}^{l}$ is the shared trainable weight matrix for all the entities at layer $l$. Weight sharing is beneficial since it is statistically and computationally more efficient than the traditional embedding methods. With the help of the weight sharing, the model can be well regularized and the number of parameters is significantly reduced.

The information propagation process of content or context can be regarded as a Markov process converges to a stationary distribution $\mathbf{P} \in \mathbb{R}^{n \times n}$ and the row $i$ indicating a likelihood of spreading from the knowledge concept $i$. This stationary distribution of the diffusion process is proven to have a closed form solution. When considering the 1-step truncation of the diffusion process, the propagation layer computes the weighted sum of the contexts' current representation. We set $\mathbf{P}^{0}$ = $\mathbf{P}^{1}$ = $\mathbf{P}^{2}$ = $\mathbf{\widetilde{D}}^{-1 / 2} \mathbf{\widetilde{A}} \mathbf{\widetilde{D}}^{-1 / 2}$,  and the three propagation layers are defined as follows:
\vspace{-5pt}
\begin{equation}
{\mathbf{h}^{1}} = ReLU(({\mathbf{P}^{0}\mathbf{X}}){\mathbf{W}^{0}}),
\end{equation}
\begin{equation}
{\mathbf{h}^{2}} = ReLU(({\mathbf{P}^{1}}{\mathbf{H}^{1}}){\mathbf{W}^{1}}),
\end{equation}
\begin{equation}
\mathbf{h}^{3} = ReLU(({\mathbf{P}^{2}}{\mathbf{H}^{2}}){\mathbf{W}^{2}}),
\end{equation}
\begin{equation}
\mathbf{e}_{MP} = {\mathbf{h}^{3}},
\end{equation}
\vspace{-2pt}
where $\mathbf{e}_{MP} \in \mathbb{R}^{d}$ is the final representations of an entity.
% The dimension $d$ of $\mathbf{e}_{MP_i}^{k}$ is further studied in experiments. $X$ is the input feature matrix of the target entities (user or knowledge concept). For example, we can utilize content feature of knowledge concepts as input to generate the representations of knowledge concepts.

%We join the three layers to each meta-path and generate the corresponding representations.
Going through the three propagation layers, we learn the representations for each meta-path. However, different meta-paths should not be considered equally. To address this problem, we utilize the attention mechanism to fuse the representation of entities learned under the guide of different meta-paths and generate the attentional joint representation. Specifically, we learn the attention weights for different meta-paths as follows:
\vspace{-5pt}
\begin{equation}
\mathbf{e}=\sum_{i=1}^{|MP|}\operatorname{att}\left(\mathbf{e}_{MP_i}\right) \mathbf{e}_{MP_i},
\end{equation}
\vspace{-2pt}
Here, $\operatorname{att}(\cdot)$ indicates the attention function. $\mathbf{e}$ indicates final representation of an entity, which has integrated the attention weights of different meta-paths. Since in this problem, we mainly focus on the user and knowledge concept. The target entity is user or knowledge concept. Formally, given the corresponding representation $\mathbf{e}_{MP_i}$ for each meta-path $MP_i \in  \{MP_1, MP_2, \cdots, MP_\mathrm{|MP|}\}$, we define the attention weights as follows:
\vspace{-5pt}
\begin{equation}
\alpha_{MP_i}=\frac{\exp \left(\sigma ( \mathbf{a}  \mathbf{e}_{MP_i}  ) \right)}{\sum_{j \in|MP|} \exp \left(\sigma ( \mathbf{a}  \mathbf{e}_{MP_j}  ) \right)},
\end{equation}
\vspace{-2pt}
where $\mathbf{e}_{MP_i}$ is the representation of an entity based on the target meta-path, and $\mathbf{e}_{MP_j}$ denotes the representation based on the other meta-paths. $\mathbf{a}$ denotes a trainable attention vector, and $\sigma$ denotes the nonlinear gating function. We formulate a feed-forward neural network to compute the correlation between one meta-path and the other meta-paths. This correlation is normalized by a softmax function. The attentional joint representation can be represented as follows:
\vspace{-5pt}
\begin{equation}
\mathbf{e}=\sum_{i=1}^{|MP|} \alpha_{MP_i} \mathbf{e}_{MP_i},
\end{equation}
\vspace{-2pt}
where $\alpha = \sum_{i=1}^{|MP|} \alpha_{MP_i}$, and $\mathbf{e}$ denotes the final representation of knowledge concepts. The meta-path attention allows us to better infer the importance of different meta-paths by leveraging their correlations and learning the entities representations.% However, we believe that a good fusion function should be learned according to the specific task. Hence, we leave the formulation and optimization of the fusion function in our recommendation model. 
The algorithm framework is shown in Algorithm \ref{alg:1}.

\begin{algorithm}[htb] 
\renewcommand{\algorithmicrequire}{\textbf{Input:}}
\renewcommand{\algorithmicensure}{\textbf{Output:}}
\caption{Generating the representations of entities.} 
\label{alg:1}
\begin{algorithmic}[1]
\REQUIRE ~~\\
the given meta-paths set $MP$; \\
the corresponding adjacency matrix set $\mathcal{A}$; \\
the features matrix $\mathbf{x}$ of target entities;\\ 
the dimension of representations $d$.\\
\ENSURE ~~\\
The representations of target entities $\mathbf{e}$.
\STATE Initialize $\mathbf{e}$;
\STATE $paths \Leftarrow []$;
\FOR{each $MP_i \in MP$}
\STATE Calculate $\mathbf{\widetilde{A}}$, $\mathbf{\widetilde{D}}$ and $\mathbf{P}$ according to Eq.1;
\STATE $\mathbf{h}^0 \Leftarrow \mathbf{x}$;
\STATE Calculate $\mathbf{h}^1$ by Eq.2;
\STATE Calculate $\mathbf{h}^2$ by Eq.3;
\STATE Calculate $\mathbf{h}^3$ by Eq.4;
\STATE Add $\mathbf{h}^3$ to $paths$;
\ENDFOR
\STATE Generate $\mathbf{e}$ by Eq.8;
\RETURN $\mathbf{e}$.
\end{algorithmic}
\end{algorithm}
\vspace{-10pt}

\subsection{Matrix Factorization for Knowledge Concept Recommendation}

So far, we have studied how to extract content feature and context feature of users and knowledge concepts, respectively. Using attention-based GCNs for representation learning, we can obtain the representation of knowledge concepts $\mathbf{e}^{k}$, and the representation of users $\mathbf{e}^{u}$.
In this part, we propose to utilize an extended matrix factorization (MF) based method to perform knowledge concept recommendation for the users. We consider the number of times users click on the knowledge concepts as a rating matrix. The rating of a user on a knowledge concept can be defined as follows:
\vspace{-5pt}
\begin{equation}
\mathbf{\widehat{r}}_{u,k}=\mathbf{x}_{u}^{\top} \mathbf{y}_{k} ,
\end{equation}
\vspace{-2pt}
where ${\mathbf{x}_{u}} \in \mathbb{R}^{D \times m}$ indicates latent factors of the user and ${\mathbf{y}_{k}} \in \mathbb{R}^{D \times n}$ denotes latent factors of the knowledge concept. $D$ is the number of latent factors. And $m$ and $n$ are the number of the user entities and knowledge concept entities. Because we have also obtained the representations for user u and knowledge concept k, we further feed them into the rating predictor as follows:
\vspace{-5pt}
\begin{equation}
\mathbf{\widehat{r}}_{u, k}=\mathbf{x}_{u}^{\top}  \mathbf{y}_{k}+\beta_{u} \cdot \mathbf{e}^{u \top} \mathbf{t}^{k}+\beta_{k} \cdot \mathbf{t}^{u \top} \mathbf{e}^{k},
\end{equation}
\vspace{-2pt}
where $\mathbf{e}^{u}$ and $\mathbf{e}^{k}$ are the representation of users and knowledge concepts. The trainable parameters $\mathbf{t}^{u}$ and $\mathbf{t}^{k}$ are introduced to make sure $\mathbf{e}^{u}$ and $\mathbf{e}^{k}$ be in the same space. $\beta_{u}$ and $\beta_{k}$ are the tuning parameters. The purpose is to achieve a suitable ratings prediction, so the object function of MF  is defined as follows:
\vspace{-5pt}
\begin{equation}
\min _{U, K}\frac{1}{m\times{n}} \sum_{u=1}^{n} \sum_{k=1}^{m} \left( \mathbf{r}_{u,k} - \mathbf{\widehat{r}}_{u,k} \right)^{2}.
\end{equation}
\vspace{-2pt}
% However, as the matter of fact, most ratings are missing. The formula can be written as following:
% \begin{equation}
% \min _{U, K}\frac{1}{2} I_{i j}\left( r_{i j} - \widehat{r_{i j}} \right)^{2}
% \end{equation}
% where the $I_{i j}$ is an function which takes the value 1 if $r_{i j}$ exists and 0 otherwise\cite{SVD}. 
We further add regularization terms to the function. 
Therefore, the final objective function is formulated as follows:
%I_{i j}
\vspace{-5pt}
\begin{equation}
\begin{split}
\min _{U, K}\frac{1}{m\times{n}} \sum_{u=1}^{n} \sum_{k=1}^{m} \left(\mathbf{r}_{u,k}-\mathbf{\widehat{r}}_{u,k}\right)^{2} + \lambda (||\mathbf{x}_{u}||_{2} \\+ ||\mathbf{y}_{k}||_{2}+||\mathbf{t}^{u}||_{2}+||\mathbf{t}^{k}||_{2}),
\end{split}
\end{equation}
\vspace{-2pt}
where $\lambda$ is the regularization parameter. We then utilize the stochastic gradient descent algorithm to optimize the local minimum of the final objective function.
%+ ||\alpha^{k}||_{2} + ||\alpha^{u}||_{2}

% $\mathrm{u}_{i}$ and $k_{j}$ are the vectors of users and knowledge concepts,  $\mathrm{b}_{i}$ and $\mathrm{b}_{j}$ are the biases of users and knowledge concepts, and
% , as follows:and the update of biases $b_{i}$, $b_{j}$ and latent factors $\mathrm{u}_{i}$, $\mathrm{k}_{j}$.
% \begin{equation}
% \begin{array}{c}{\mathrm{b}_{i} \leftarrow b_{i}+\gamma\left(\mathrm{e}_{i j}-\lambda \mathrm{b}_{i}\right)} \\ {\mathrm{b}_{j} \leftarrow b_{j}+\gamma\left(e_{i j}-\lambda b_{j}\right)} \\ {\mathrm{u}_{i} \leftarrow u_{i}+\gamma\left(\mathrm{e}_{i j} \cdot k_{j}-\lambda \mathrm{u}_{i}\right)} \\ {\mathrm{k}_{j} \leftarrow k_{j}+\gamma\left(\mathrm{e}_{i j} \cdot u_{i}-\lambda \mathrm{k}_{j}\right)}\end{array}
% \end{equation}
% where $e_{i j} = r_{i j} - \mathrm{u_{i}^{T}}\mathrm{k_{j}}$ and $\lambda$ is the learning rate. The derivations can be calculated in the above manner for both users and items. 
% As shown in Algorithm 2, it is the overall algorithm framework of our model.
 
% \subsection{Model Training}

% The two models of the profile reviser and the basic recommendation model are interleaved together, and we need to train them jointly. The training process is shown in Algorithm 1, where we firstly pre-train the basic recommendation model based on the original dataset, then we fix the parameters of the basic recommendation model and pre-train the profile reviser to automatically revise the user profiles; finally, we jointly train the models together.
\vspace{-5pt}
\section{EXPERIMENTS}\label{sec:exper}

\subsection{Datasets}

We collected real world data from $XuetangX$ MOOC platform. {We select enrollment behaviors occurring between October 1st, 2016 and December 30th, 2017 as the training set, and those occurring between January 1st, 2018 and March 31st, 2018 as the test set.} Each instance in the training set or test set is a sequence representing a user's history of click behaviors. In the training process, for each sequence in the training data, we treat the last clicked knowledge concept as the target and the remainders as the past behaviors. Moreover, for each positive instance, we randomly generate one negative instance to replace the target knowledge concept. In the testing process, we treat each enrolled knowledge concept in the test set as the target knowledge concept; the corresponding knowledge concepts of the same user in the training set are treated as the sequence representing the history of clicked knowledge concepts. To evaluate the recommendation performance, each positive instance in the test set is paired with 99 randomly sampled negative instances, and outputs prediction scores for the 100 instances (1 positive and 99 negatives)\cite{NAIS}.

\subsection{Evaluation Metrics}
We evaluate all the methods in terms of the widely used metrics, including Hit Ratio of top-K items (\textit{HR@K}) and Normalized Discounted Cumulative Gain of top-K items (\textit{NDCG@K}) \cite{NDCG}. HR@K is a recall-based metric that measures the percentage of ground truth instances that are successfully recommended in the top K, and \textit{NDCG@K} is a precision-based metric that accounts for the predicted position of the ground truth instance. We set K to 5, 10, and 20, and calculate all metrics for every 100 instances (1 positive plus 99 negatives). The final recommendation list for user $u$ is $R_{u}=\left\{r_{u}^{1}, r_{u}^{2}, \cdots, r_{u}^{K}\right\}$. $r_u^i$ denotes rank at the $i-th$ position in $R_u$ based on the predict score. $T_u$ is the interacted items set of user $u$ in the test data, and $N$ is the total number of users in our test data.
% and report the average score of all users.
\vspace{-5pt}
\begin{equation}
HR@K=\frac{1}{N}\sum_{u}I(|R_u \cap T_i|) ,
\end{equation}
\vspace{-2pt}
where, $I(x)$ is an indicator function whose value is 1 when $x > 0$ and $0$ otherwise. The large the value of $HR@K$ the better the performance of the model.
\vspace{-5pt}
\begin{equation}
NDCG@K = \frac{1}{Z}DCG@K = \frac{1}{Z}\sum_{j=1}^{K} \frac{2^{I(|\{r_u^j\} \cap T_u|)}-1}{log_2(j+1)} ,
\end{equation}
\vspace{-2pt}
where $Z$ is a normalization constant which is the maximum possible value of $DCG@K$. We also use the mean reciprocal rank (\textit{MRR}). From the definition, we can see that a larger \textit{MRR} value indicates a better performance of the model\cite{NAIS}.
\vspace{-5pt}
\begin{equation}
\mathrm{MRR}=\frac{1}{N} \sum_{i=1}^{N} \frac{1}{\operatorname{rank}_{i}} ,
\end{equation}
\vspace{-2pt}
where $rank_{i}$ refers to the rank position of the one positive instance for the $i-th$ user in 100 instances. In addition, we also add the area under the curve of ROC (\textit{AUC}) as a metric.

\subsection{Detailed Analysis of the Proposed Approach}

\begin{table}
    \caption{Different results from different combinations of meta-paths.}
    \vspace{-5pt}
    \label{tab:metapaths}
    \begin{tabular}{cccccl}
      \toprule
      Meta-path & HR@5 & NDCG@5 & MRR & AUC \\
      \midrule
      MP1 & 0.5393 & 0.3817 & 0.3621 & 0.8645 \\
      MP2 & 0.4508 & 0.3136 & 0.3059 & 0.8487 \\
      MP3  & 0.5870 & 0.4215 & 0.3972 & 0.8796 \\
      MP4 & 0.4302 & 0.3016 & 0.2967 & 0.8314 \\
      MP1 \& MP2 & 0.5669 & 0.3962 & 0.3749 & 0.8824 \\
      MP1 \& MP3 & 0.6114 & 0.4295 & 0.4042 & 0.9091 \\
      MP1 \& MP4 & 0.5936 & 0.4157 & 0.3891 & 0.8899 \\
      MP2 \& MP3 & 0.6062 & 0.4273 & 0.3998 & 0.8927 \\
      MP2 \& MP4 & 0.4541 & 0.3210 & 0.3151 & 0.8469 \\
      MP3 \& MP4 & 0.6011 & 0.4233 & 0.3950 & 0.8871 \\
      MP1 \& MP2 \& MP3 & 0.6404 & 0.4543 & 0.4240 & 0.9115 \\
      MP1 \& MP2 \& MP4 & 0.6029 & 0.4279 & 0.4022 & 0.8955 \\
      MP1 \& MP3 \& MP4 & 0.6212 & 0.4389 & 0.4102 & 0.9021 \\
      MP2 \& MP3 \& MP4 & 0.6025 & 0.4318 & 0.4097 & 0.8932 \\
      MP1\&MP2\&MP3\&MP4 & \textbf{0.6470} & \textbf{0.4635} & \textbf{0.4352} & \textbf{0.9232} \\
      \bottomrule
    \end{tabular}
    \vspace{-8pt}
\end{table}
\vspace{-5pt}
\subsubsection{Evaluation of Different Meta-paths Combination}

In this part of the experiments, we analyze how selection of meta-path combinations affect the performance of $\method$, since a small number of high-quality meta-paths can lead to considerable performance \cite{Heterogeneous}. We consider both single meta-path and their combinations. 
Specifically, We select four types of meta-paths to characterize the relatedness between pair of users, including \textbf{$MP_1$}: $U \rightarrow K \stackrel{\text { -1 }}{\longrightarrow} U$, \textbf{$MP_2$}: $U \rightarrow C \stackrel{\text { -1 }}{\longrightarrow} U$, \textbf{$MP_3$}: $U \rightarrow V \stackrel{\text { -1 }}{\longrightarrow} U$, and \textbf{$MP_4$}: $U \rightarrow C \rightarrow T \stackrel{\text { -1 }}{\longrightarrow} C \stackrel{\text { -1 }}{\longrightarrow} U$. 
We also select three types of meta-path to characterize the relatedness between pair of knowledge concepts, including 
$K \leftrightarrow K$, $K \rightarrow U \stackrel{\text { -1 }}{\longrightarrow} k$ and $K \rightarrow C \stackrel{\text { -1 }}{\longrightarrow} K$. 
To analyze the impact of different combinations in a small number of meta-paths, we use all three meta-paths to model the knowledge concept and study the performance with single user related meta-path and their combinations. The experiments results are shown in Table \ref{tab:metapaths}. 
From Table \ref{tab:metapaths}, we can find that each single meta-path (i.e., \textbf{$MP_1$, $MP_2$, $MP_3$, $MP_4$}) exhibits different performance, where the performance ranking is \textbf{$MP_3$} > \textbf{$MP_1$} > \textbf{$MP_2$} > \textbf{$MP_4$}, and the combinations of single meta-paths follow the same tendency (e.g., the performance of \textbf{$MP_1\&MP_3$} > \textbf{$MP_1$\&$MP_2$}, \textbf{$MP_1$\&$MP_2$\&$MP_3$} > \textbf{$MP_1\&MP_2\&MP_4$}). This illustrates that different meta-paths indicate different relations
%and that different meta-paths have different adaptive weights, which are trained in the attention mechanism. 
Further, although the growth of performance is not quite obvious, we can observe that the combination including more meta-paths will exhibit better performance, and the best performance is achieved by combining all four meta-paths. 

\subsubsection{Evaluation of Model Parameters.}

\begin{figure*}
\centering
\includegraphics[width=\textwidth]{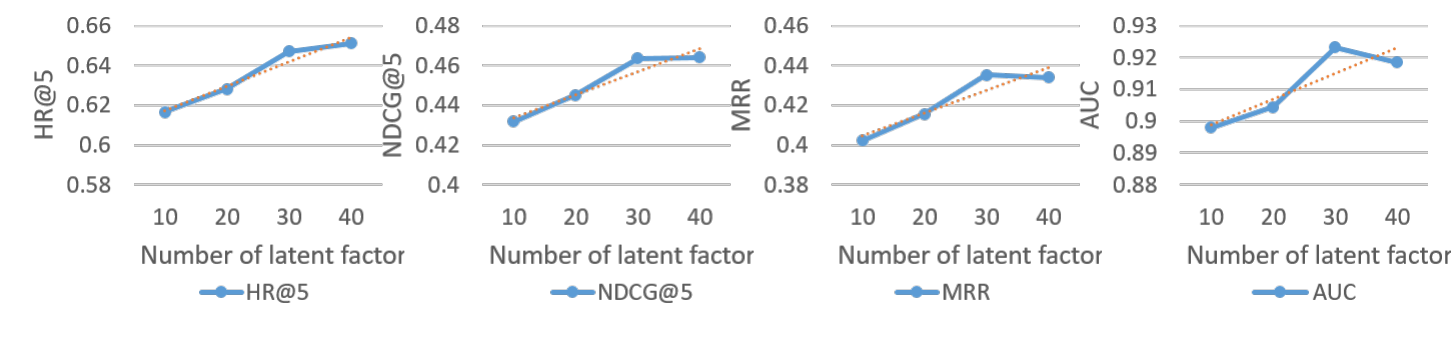}  
\vspace{-28pt}
\caption{Performance of different number of latent factors $D$.}
\label{fig:latent_number}
\vspace{-10pt}
\end{figure*}

In a matrix factorization-based method, the number of latent factors is an important parameter. Therefore, we present a comparison of the performance obtained with different numbers of latent factors. As shown in Figure \ref{fig:latent_number}, we select the metrics $HR@K$, $NDCG@K$, $MRR$, and $AUC$ to show how the performance of $\method$ changes with changes in the number of latent factors. We tune the number from 10 to 40 in increments of 10. We can see the increase in performance becomes flat as the number of latent factors increases. We find that using 30 latent factors can produce optimal performance.

\begin{figure*}
\centering
\includegraphics[width=\textwidth]{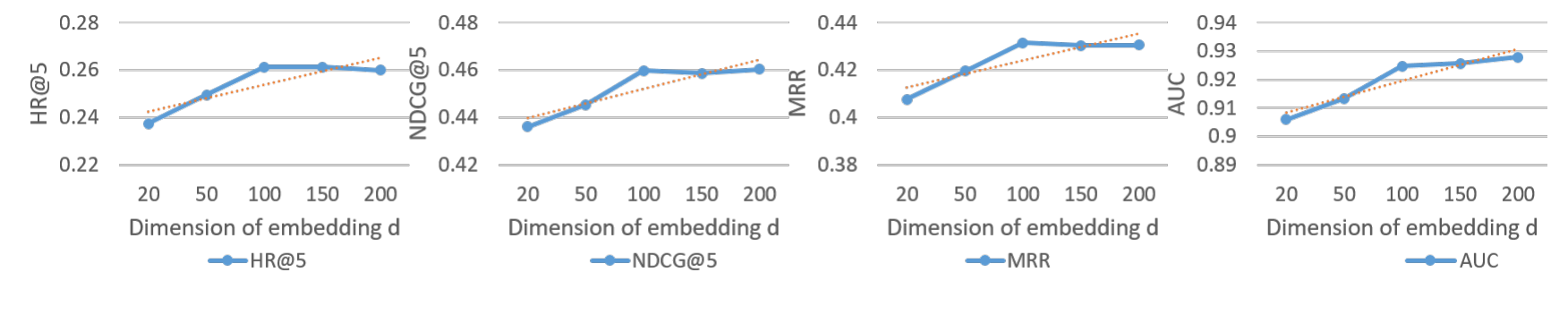} 
\vspace{-27pt}
\caption{Performance of different dimension of representations $d$.}
\label{fig:embedding}
\vspace{-5pt}
\end{figure*}

After setting the number of latent factors as 30, we study the dimension settings of the entities representation. The experiment results are shown in Figure \ref{fig:embedding}. We conducted the experiments using different numbers of dimensions (i.e., 20, 50, 100, 150, 200), and found that optimal performance was achieved with 100. Therefore, both the user and the knowledge concepts are represented as 100-dimensional vectors. The results also show that the representation of user and knowledge concepts in the heterogeneous information network is an important factor in improving the performance of the recommendation task. 
%We can easily find that the representation dimension is an important parameter. The representation dimension impacts effectively on the final performance of the model.

\begin{figure}
\centering
\includegraphics[width=8cm]{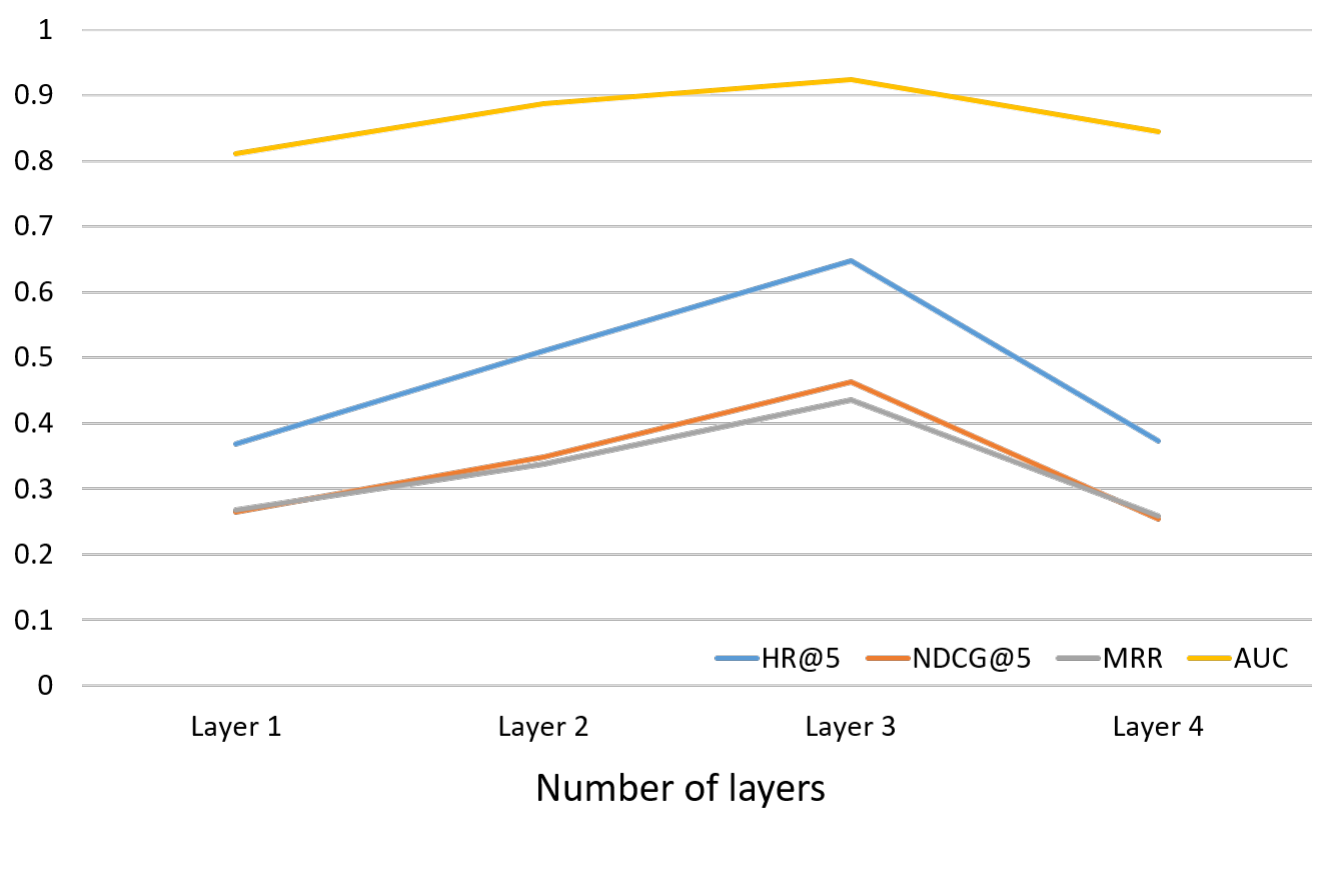}  
\vspace{-9pt}
\caption{Performance of different number of layers.}
\label{fig:layers}
\vspace{-15pt}
\end{figure}

We also examine how the number of GCN layers influence the performance of the model. As shown in Figure \ref{fig:layers}, we can clearly see that the performance of the proposed model changes with different numbers of layers (i.e., 1, 2, 3, and 4). It illustrates that the optimal number of GCN layers is around 3. 
%What's more, we also show that the importance of layers in neural networks.

\vspace{-5pt}
\subsection{Baseline Methods}

\begin{table*}
  \caption{Results obtained with different models using the MOOC dataset.}
  %\vspace{-5pt}
\label{tab:Results}
\begin{tabular}{ccccccccccl}
  \toprule
   & HR@1 & HR@5 & HR@10 & HR@20 & NDCG@5 & NDCG@10 & NDCG@20 & MRR & AUC \\
  \midrule
  $BPR$ & 0.1699 & 0.4633 & 0.6246 & 0.7966 & 0.3217 & 0.3736 & 0.4151 & 0.3156 & 0.8610 \\
  $MLP$ & 0.0660 & 0.3680 & 0.5899 & 0.7237 & 0.2231 & 0.2926 & 0.3441 & 0.2146 & 0.8595 \\
  $FM$ & 0.2272 & 0.4057 & 0.5867 & 0.7644 & 0.3655 & 0.3968 & 0.3930 & 0.3067 & 0.8574 \\
  $FISM$ & 0.1410 & 0.5849 & 0.7489 & 0.7610 & 0.3760 & 0.4203 & 0.4279 & 0.3293 & 0.8532 \\
  $NAIS$ & 0.078 & 0.4112 & 0.6624 & 0.8649 & 0.2392 & 0.3201 & 0.3793 & 0.2392 & 0.8863 \\
  $NASR$ & 0.1382 & 0.4437 & 0.6215 & 0.7475 & 0.2364 & 0.3172 & 0.3821 & 0.2117 & 0.8215 \\
  $metapath2vec$ & 0.2476 & 0.5983 & 0.7598 & 0.8689 & 0.4194 & 0.4422 & 0.4602 & 0.3873 & 0.8909 \\
  \hline
  \hline
  $ACKRec_{h}$ & 0.2092 & 0.5388 & 0.7139 & 0.8665 & 0.3783 & 0.4348 & 0.4738 & 0.3634 & 0.8927 \\
  $ACKRec_{c}$ & 0.2457 & 0.5917 & 0.7542 & 0.8778 & 0.4216 & 0.4763 & 0.5079 & 0.4026 & 0.8974 \\
  $ACKRec_{s}$ & 0.2195 & 0.5917 & 0.7476 & 0.8553 & 0.4154 & 0.4659 & 0.4933 & 0.3891 & 0.8848 \\
  $ACKRec_{r}$ & 0.2588 & 0.6427 & 0.7911 & 0.8909 & 0.4591 & 0.5074 & 0.5329 & 0.4285 & 0.9035 \\
  $ACKRec_{s+r}$ & \textbf{0.2645} & \textbf{0.6470} & \textbf{0.8122} & \textbf{0.9255} & \textbf{0.4635} & \textbf{0.5170} & \textbf{0.5459} & \textbf{0.4352} & \textbf{0.9232} \\
  \bottomrule
\end{tabular}
\end{table*}

To evaluate the performance of the proposed approach, we consider various of baseline methods as follows:
\begin{itemize}
  \item $BPR$ \cite{BPR}: It optimizes a pairwise ranking loss for the recommendation task in a Bayesian manner.
  \item $MLP$ \cite{MLP}: It applies a multi-layer perceptron (MLP) to a pair of user representations and corresponding knowledge concept representations to learn the probability of recommending a knowledge concept to the user.
  \item $FM$ \cite{FM}: This is a principled approach that can easily incorporate any heuristic features. However, to ensure a fair comparison to other methods, we only use the representations of users and knowledge concepts.
  \item $FISM$ \cite{FISM}: This is an item-to-item collaborative filtering algorithm that conducts recommendations based on the average embeddings of all behavior histories and the embeddings of the target knowledge concept.
  \item $NAIS$ \cite{NAIS}: This is also an item-to-item collaborative filtering algorithm, but distinguishes the weights of different online learning behaviors using an attention mechanism method.
  \item $NASR$ \cite{NASR}: This is an improved GRU \cite{hidasi2015session} model that estimates an attention coefficient for each behavior history based on the corresponding hidden vector output by GRU. And GRU is a gated recurrent unit model which receives a list of historical behavior as input.
  \item $metapath2vec$ \cite{metapath2vec}: This is a meta-paths based representation method in heterogeneous information network by random walk and skip-gram. With the same meta-path of user and knowledge concept in \textbf{ACKRec} model, we use metapath2vec to generate the representations of users and knowledge concepts with the same dimension of \textbf{ACKRec}.
%   Then we feed them into the matrix factorization model.
  \item $ACKRec_{h}$: A variant of \method, which ignores the heterogeneity of entities in the heterogeneous information network. We regenerate the adjacency matrix to represent the relationship between different entities.
  \item $ACKRec_{c}$: This can be viewed as a variant of \method without the attention mechanism method, and concatenates different meta-paths together.
  \item $ACKRec_{s}$: Content feature-based \method. The input of this model is just the content feature of entities.
  \item $ACKRec_{r}$: Context feature-based \method.  Similar to the model $ACKRec_{s}$, the context feature of entities in MOOCs are fed into this model.
  \item $ACKRec_{s+r}$: The proposed method, which combines heterogeneous context feature and content feature of entities to maximally depict the entities in the HIN.
\end{itemize}
\vspace{-5pt}

For the MOOCs dataset, we split the user history behavior data into a training set and a test set. The training time for our method and the baseline methods are followed: \textbf{BPR} 2hrs, \textbf{MLP} 2hrs, \textbf{FM} 2.5hrs, \textbf{FISM} 4hrs, \textbf{NAIS} 3.5 hrs, \textbf{NASR} 4.5 hrs, \textbf{metapath2vec} 5.5 hrs, \textbf{ACKRec} (our method) 4.5 hrs. 
%and the response time is about 20ms per sample. 
As shown in Table \ref{tab:Results}, we compare \method with other machine learning methods.  For the HIN based methods, We select meta-paths combination given the best performance in Section 3.1.2. As shown in Table \ref{tab:Results}, HIN based methods outperform all the other methods. It indicates the importance of the heterogeneity in MOOCs data
%that real world data include multiple types of entities and we hence should not ignore the important heterogeneity of objects. 
Furthermore, the results show that $ACKRec_{s+r}$, which integrates content feature and context feature, gives the best performance. 
%significantly outperforms the other seven baselines.
Different from metapath2vec \cite{metapath2vec} generating representations based on random walk strategy and skip-gram method to generate representations of nodes, our model utilizes the graph convolutional networks to learn representations and adaptive attention mechanism to learn the different meta-paths weights and can better capture the heterogeneity within the data. 

In addition, compared with $ACKRec_{h}$, it obviously demonstrates that \method utilizing the meta-paths based method on the HIN can more effectively capture the heterogeneous relationships. Compared with $ACKRec_{c}$, we can see that adaptively fuse representation learned in different meta-path is better than simple concatenation, since different meta-path has different importance corrsponds to the task.
%different meta-paths have unequal weights for the representation of entities. The attention mechanism-based method is able to dynamically assign weights to different meta-paths. 
Moreover, compared with $ACKRec_{s}$ and $ACKRec_{r}$, it demonstrates that methods utilizing either content feature or context feature alone will lose information needed for the representation of entities; thus, they cannot comprehensively depict the feature of users and knowledge concepts in MOOCs. The final method $ACKRec_{s+r}$, which uses adaptive weights and the meta-path based approach over the HIN, can integrate rich content feature of entities and structural relations between different types of entities in a more comprehensive and effective manner, and achieves the best performance.

\vspace{-5pt}
\subsection{Case Study}
\begin{figure}
\centering
\includegraphics[width=8.5cm]{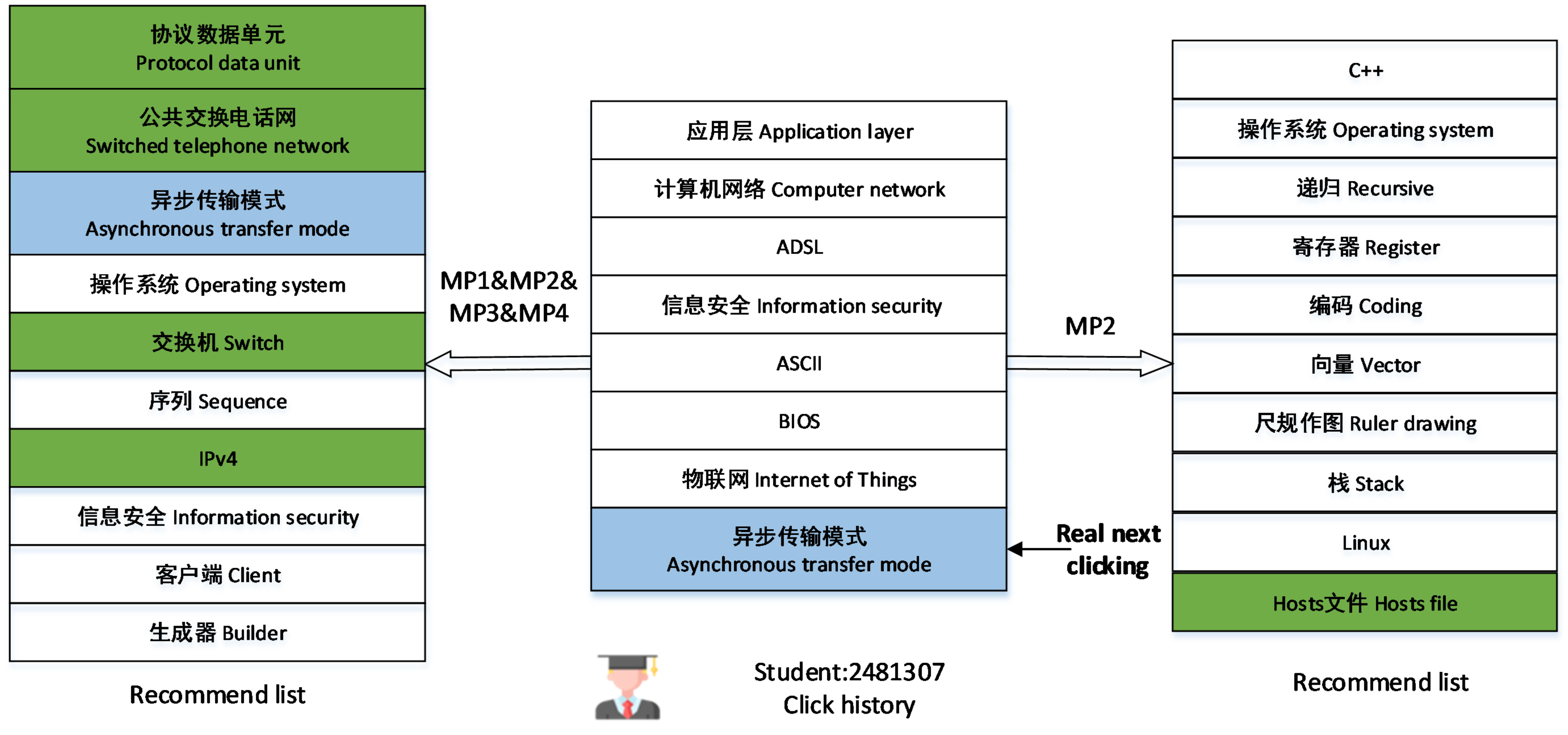}  
\vspace{-7pt}
\caption{The case study of \method bases different meta-paths. The blue labels denote the real next click, the green labels are the related knowledge concepts of the field of user \textit{student:2481307} interests in. The left set is the recommendation list which bases meta-paths \textbf{$MP_1-MP_2-MP_3-MP_4$}; the middle set is the leaning history of \textit{student:2481307}; the right set is the recommendation list which bases a single meta-path \textbf{$MP_2$}.}
\label{fig:case}
\vspace{-18pt}
\end{figure}

In this part, we conduct one case to demonstrate the effectiveness of our proposed method \method. We randomly select a \textit{student:2481307} and obtain two top 10 recommend lists based on single meta-path \textbf{$MP_2$} and combined meta-paths \textbf{$MP_1-MP_2-MP_3-MP_4$}, respectively. As shown in Figure \ref{fig:case}, we can intuitively observe that the proposed model generates different results with different reality conditions. With more meaningful relationships among user and knowledge concept, the recommend list will contain more related knowledge concepts for the corresponding user. For example, from the click history list of $student:2481307$, we know that the student is especially interested in the field of computer network, the result of meta-paths \textbf{$MP_1-MP_2-MP_3-MP_4$} shows that the \method can capture the student's interests of knowledge concept, matching his real next click, \textit{asynchronous transfer mode}, shown in blue.  The other recommendations such as \textit{protocol data unit}, \textit{switched telephone network},  \textit{switch}, and \textit{IPv4} are also highly related. In particularly, the recommended results will be significantly different for different meta-paths.

% Secondly, as is shown in Figure \ref{fig:case2}, the $student:647984$ received different recommend list $a$ and recommend list $b$. In fact, both of them hit the real next click. There are four identical knowledge concepts with different indexes in the two recommendation lists. In additionally, we can find that our model \method will generate different results which base the disparity clicking history of the same user. This case demonstrates that the flexibility of our model \method, and the graph convolutional networks is able to dynamically extract the structure of the entity in a heterogeneous information network. This ability efficaciously improve the effect of knowledge concept recommendation in MOOCs.

% \begin{figure}
% \centering
% \includegraphics[width=8cm]{images/case_study4.pdf}  
% \vspace{-5pt}
% \caption{The case study of \method in a $student:647984$ with different click behavior. The different of two click history list is the last behavior. The blue labels denote the real next clicking. Both of two recommend list base meta-paths $MP_1-MP_2-MP_3-MP_4$ and hit the real next clicking. We can find that the orange labels in two recommend list is different from each other.}
% %  
% \label{fig:case2}
% \vspace{-12pt}
% \end{figure}

% 图片粘贴不上

\vspace{-5pt}
\section{Related Work}\label{sec:relatedwork}

\subsection{Graph Neural Network in Heterogeneous Information Network}
Graphs play a crucial role in modern machine learning\cite{Representation,Anewmodel}. Recently, graph neural networks\cite{GraphNeural, kipf2016semi,Session-based,GraphConvolutional,Adaptive,FastGCN:,ConvolutionalNeural} have become recurrent topics in machine learning, and both have broad applicability. However, in the real world, the graphs are usually heterogeneous. There are a few attempts heterogeneous information network setting.
%Peng et al. \cite{peng2019fine} proposed PP-GCN based fine-grained social event categorization model in an event-based HIN for mining social events. 
% Wu et al.\cite{Session-based} constructed structured graph data using session sequences in the recommendation. The PinSage\cite{GraphConvolutional} model combined random walks and graph convolutions to generate node embeddings that incorporate both graph structure and node feature information. In order to learn graphics effectively, Li et al.\cite{Adaptive} proposed adaptive graph convolutional neural networks. Chen et al.\cite{FastGCN:} performed the FastGCN model, which is based on the Monte Carlo approach, to reduce the computational footprint. In particular, Gama et al.\cite{ConvolutionalNeural} verified the GNN performance of different architectures. 
% For the heterogeneous information network, there are also some attempts.
Wang et al. \cite{wang2019attentional} proposed DeepHGNN, an attentional heterogeneous graph neural network model to learn from the heterogeneous program behavior graph to guide the reidentification process. 
%The learn the heterogeneous graph representation with considering the node-wise, layer-wise, and path-wise context importance, 
Wang et al. \cite{wang2019heterogeneous} presented HAGNN, a Hierarchical Attentional Graph Neural Encoder and used it for program behavior graph analysis.
%In the intrusion detection field, Wang et al.\cite{wang2019attentional} presented DeepRe-ID, a model based on an attentional multi-channel graph neural network, to identify the identity of programs. 
Additionally, the GEM\cite{HeterogeneousGraph} model, a heterogeneous graph neural network approach for detecting malicious accounts at Alipay, has been presented. Unlike these approaches, our proposed model utilizes attentional graph convolutional networks for the representations of users and knowledge concepts in heterogeneous information networks.
% Unlike these approaches, our proposed model utilizes attention-based GCN for the representations of users and knowledge concepts in heterogeneous information networks.
\vspace{-8pt}
\subsection{Recommendation System in Heterogeneous Information Network}

Some information recommendation models are based on heterogeneous information networks. 
\cite{AGeneral} proposed Heaters, a graph-based model, to solve the general recommendation problem in heterogeneous networks.
%\cite{Recommendation} pointed out that the main challenge in building recommendation systems in HINs involves defining the characteristics of the system to represent different types of relationships between entities and to learn the importance of each type of relationship.
Yu et al.\cite{Personalized} proposed to use meta-paths based latent features to represent the connectivity between users and items along with different types of paths. Additionally,  
%For the interaction of events, Gui et al.\cite{EmbeddingLearning} proposed a framework based on hyperedge embedding. 
Follow precious work, Shi et al. \cite{Semantic,hu2018leveraging,Heterogeneous} proposed to use meta-path concept to mode the heterogeneous information in HIN.
Different from previous methods, this study focuses on capture the representations of different types of entities on the heterogeneous information network and fuses themselves content feature of different types of entities and the structure features of entities in MOOCs data together for the recommendation task of the knowledge concept.
\vspace{-10pt}
\section{CONCLUSION}\label{sec:conclu}
In this work, we investigate the problem of the knowledge concept recommendation in MOOCs system, which is often overlooked by MOOCs recommendation system. We propose \method, an end-to-end graph neural network based approach that naturally incorporates rich heterogeneous context side information into knowledge concept recommendation. To make use of rich context information in a more natural and intuitive way, we model the MOOCs as a heterogeneous information network. We design an attention-based graph convolutional network to learn the representation of different entities via propagate context information under the guide of meta-path in an attentional way. With the help of proposed attention-based graph convolutional network, the users' potential interests can be effectively explored and aggregated. 
Comprehensive experimental study on real data collected from XuetangX is conducted. The proposed approaches outperform the strong baseline. 
The promising experimental results illustrate the effectiveness of the proposed method. 
\vspace{-5pt}
\section{Acknowledgments}
%This work is supported , and NSF under grants III-1526499, III-1763325, III-1909323 and CNS-1930941. The authors also would like to thank XuetangX for data collection and supports.
This work is supported by NSF under grants III-1526499, III-1763325, III-1909323, CNS-1930941, by Science and Technology Project of the Headquarters of State Grid co., LTD under Grant No. 5700-202055267A-0-0-0, and by NKPs under grants 2018YFC0830804.
%in part by Science and Technology Project of the Headquarters of State Grid co., LTD under Grant No. 5700-202055267A-0-0-0, and the National Key R&D Program of China under grant 2018YFC0830804. 
The authors also would like to thank \textit{\textbf{XuetangX}} for data collection and supports.
\balance
\bibliographystyle{ACM-Reference-Format}
\bibliography{sample-base}

%%
\end{document}